\documentclass[review]{elsarticle}

\usepackage{lineno,hyperref}
\modulolinenumbers[5]
\usepackage{amsmath}
\usepackage{multirow}

\journal{Pattern Recognition}

\begin{document}

\begin{frontmatter}

\title{Pose-driven Attention-guided Image Generation for Person Re-Identification}





\author{Amena Khatun\corref{cor1}} 
\cortext[cor1]{Corresponding author: 
  Tel.: +61414186468;}  
 % fax: +0-000-000-0000;}
\ead{a2.khatun@qut.edu.au}
\author{Simon Denman}
\author{Sridha Sridharan}
\author{Clinton Fookes}

\address{Signal Processing, Artificial Intelligence and Vision Technologies (SAIVT), Queensland University of Technology (QUT), Brisbane, QLD 4000, Australia}

\begin{abstract}
Person re-identification (re-ID) concerns the matching of subject images across different camera views in a multi camera surveillance system. One of the major challenges in person re-ID is pose variations across the camera network, which significantly affects the appearance of a person. Existing development data lack adequate pose variations to carry out effective training of person re-ID systems. To solve this issue, in this paper we propose an end-to-end pose-driven attention-guided generative adversarial network, to generate multiple poses of a person. We propose to attentively learn and transfer the subject pose through an attention mechanism. A semantic-consistency loss is proposed to preserve the semantic information of the person during pose transfer. To ensure fine image details are realistic after pose translation, an appearance discriminator is used while a pose discriminator is used to ensure the pose of the transferred images will exactly be the same as the target pose. We show that by incorporating the proposed approach in a person re-identification framework, realistic pose transferred images and state-of-the-art re-identification results can be achieved.
\end{abstract}


\end{frontmatter}

%\linenumbers

\section{Introduction}
\label{sec:intro}
\begin{figure}
\begin{center}
\includegraphics[width=0.85\linewidth]{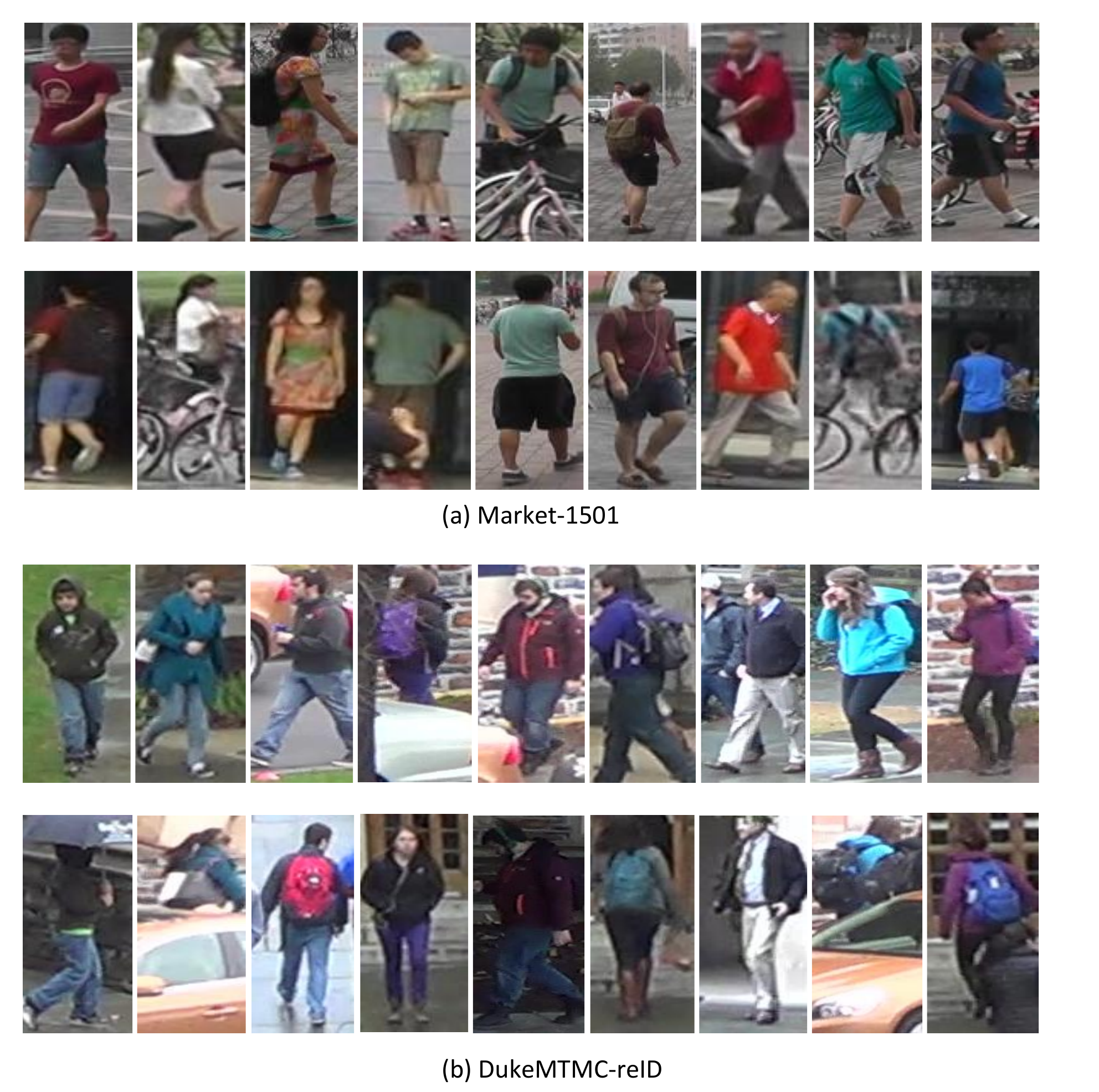}
\end{center}
\caption{Illustration of pose variations in Market1501 and DukeMTMC-reID. Camera viewpoint and pose variations can significantly affect the appearance of a person. For instance, the first person from Market-1501 is wearing a red t-shirt and carrying a backpack that is only visible in one camera view. In the second column, the lady wearing black skirt with a white top appears completely different in another camera due to occlusion caused by bicycles. }
\label{fig:Dataset_comparison_ch7}
\end{figure}

Variations in human pose is one of the key challenges in person re-identification (re-ID), as subject pose and camera angles vary between cameras and over time. The appearance of a person can significantly change between observations due to pose variations and this has a negative impact on person re-ID. For instance, the same person may be seen to carry a backpack in one camera when a side or rear view is available, yet in a front view captured by a different camera as the person move through the camera network, the backpack may not be visible. Variations in human pose across camera views are illustrated in  Figure \ref{fig:Dataset_comparison_ch7}. While some researchers introduced hand-crafted methods \cite{DPFL,Liao_2015_CVPR,8237527,8099654,6619307,DNS} to address challenges due to pose variations, others developed deep learning based frameworks and divided the human body into several parts \cite{su2017pose,zhao2017spindle,Zhao_2017_ICCV}, extracting features from individual body parts or regions, arguing that this body part partitioning is helpful for improving robustness to pose variation.

Other studies \cite{ge2018fd,ZHANG202022,Liu_2018_CVPR,10.1007/978-3-030-01240-3_40,saquib2018pose} adopted generative adversarial networks (GAN) to generate new poses of a person. Existing GAN based pose transfer re-ID methods used either a GAN \cite{NIPS2014_5423} or conditional GAN \cite{DBLP:journals/corr/MirzaO14} for pose-guided image generation. However, the appearance of the same individual can be drastically different, particularly if the camera can capture only a partial view of the person, which makes it challenging to transfer a person's pose. For example, compared to transferring the pose of a person from a front view to a side view, it is much more difficult to generate a realistic pose transformed image when transferring from a person riding a bicycle to a standing position. In these situations, existing GAN based methods fail to generate realistic images as they transfer only the global structure from the target pose to the new image. As such, it is essential to learn the local details of a person alongside the global structure to smoothly transfer between poses.

To tackle these issues, in this paper we propose a pose-guided person image generation network to sequentially transfer a subject from their original pose to a pose defined by a condition image. To do this we use two sub-networks: a pose attention-guided appearance network, and a pose attention-guided generation network. Each network is composed of a series of blocks, and each pose appearance block captures the local features within the global structure. For image generation, it is essential to extract and separate the components that contribute to appearance, local details and pose, and to correctly combine them back in to a person image. 
More specifically, the pose transfer network consists of an appearance encoder and a pose encoder, followed by a sequence of pose attention-guided appearance modules to better select the regions of a human body that need to be transferred. The modules sequentially learn and update the appearance and pose representations of the person. In the image generator, the same attention mechanism is applied as is used in the pose attention-guided appearance modules to recombine the appearance and pose representations to generate high quality pose-transferred images. 

In \cite{Khatun_2020_WACV}, we proposed a semantic-consistency loss to help preserve semantic details between the source and transferred images from the same identity. Motivated by the success of this mechanism, we adopted our earlier proposed semantic-consistency loss between the condition image and the pose-transferred image to ensure the preservation of semantic information during pose translation. The newly generated pose-transferred images help to enhance the discriminative ability of the re-identification model, and thus increase re-ID performance. For re-ID, we adopted our earlier proposed improved quartet loss for verification \cite{KHATUN2020102989} and softmax identification loss. The major contributions of this paper can be summarised as follows:

\begin{itemize}
    \item We propose a novel identity-preserving pose transfer network to transfer the pose of a person in a source image to a target pose without affecting the identity of the person, allowing a re-ID framework to learn from a more diverse set of poses than is otherwise possible.
    \item We propose a method to sequentially transfer an image to a target pose by a series of pose-guided attention modules. Thus each module gradually learns and updates the appearance and pose representations of a person, allowing them to capture local features alongside the global structure.
    \item Our approach preserves the identity of the person in the pose translated images by using the semantic-consistency loss. An appearance discriminator is used to ensure that the fine image details are realistic after pose translation while a pose discriminator is introduced to ensure the target image correctly captures the shape defined by the target pose.
\end{itemize}

\section{Related Work}

\subsection{Generative Adversarial Networks}
Generative adversarial networks (GANs) were first introduced in \cite{NIPS2014_5423} to generate synthetic images which follow the same distribution as a corpus of real images. While GANs use noise alone as an input and thus offer no control over the generated data, the conditional GAN (cGAN) \cite{8100115} regulates the mode of the  generated images by using two inputs: a noise sample and a conditional input. However cGANs \cite{8100115} require paired data for training which can be challenging to obtain. To address this, CycleGAN is introduced in \cite{CycleGAN2017} to transfer the style of an image between two domains, and through the use of a cycle consistency loss, it does not require paired training data. A similar approach is proposed in \cite{Yi2017DualGANUD} and uses dual learning to train a translator network with two sets of images from two domains, thus avoiding the need for paired data during training. Motivated by the success of GANs, many studies focus on various applications of GAN such as image translation \cite{tang2019cycleincycle,7780634,8658643} between domains, pixel-level transfer from source to target domains \cite{Yoo2016PixelLevelDT}, style transfer between domains \cite{Yi2017DualGANUD}, and pose-transfer from source to target pose \cite{8578457,tang2020bipartite, tang2020xinggan,zhu2019progressive}.

\subsection{Deep Learning for re-ID}
Since deep convolutional neural networks (DCNNs) integrate the feature extraction and metric learning together, DCNNs are widely adopted by person re-ID researchers to achieve improved performance. The works in \cite{DBLP:journals/corr/VariorHW16,7780513,6909421} developed Siamese networks for re-ID where the model is trained using a pair of input images to pull images from the same person close to each other. Other researchers have adopted a triplet framework that reduces the distance between a positive pair more than the distance between a negative pair in feature space, with respect to the same probe image. The triplet loss function is improved in \cite{Cheng_2016_CVPR} by further minimising the distance between the positive pair by adding a new threshold. In \cite{DBLP:journals/corr/ChenCZH17}, the triplet loss is improved using two margins, the first margin is performing the same as the triplet loss and the second is attempting to maximise inter-class distance. The second margin is however weaker than the first, which leads the network to being dominated by the conventional triplet loss, i.e. minimising the intra-class distance when the probe images
belong to the same identity. This issue is addressed in \cite{Amena} by a new loss function that minimises the intra-class distance more than the inter-class distance with respect to multiple different probe images. However, images from the same identity may lie close to each other in feature space, but with a large intra-class distance as \cite{Amena} does not specify the distance between the positive pair. Thus, \cite{KHATUN2020102989} proposed a new loss function that uses a quartet of images and minimises the distance between the positive pair more than the distance between the negative pair regardless of whether the probe image comes from the same identity or not, and simultaneously ensures that the intra-class features will be close to each other.

To overcome occlusions and  challenges caused by partial observation across camera networks, some recent re-ID approaches \cite{Li_2017_CVPR,zhao2017spindle,98a1e05749b24099a51dcf3c22daefd9, Yang2019PatchBasedDF} proposed part-based methods. These methods, however, all rely on verification or identification frameworks only instead of jointly adopting both approaches, and some methods require additional supervision and extra annotation. Other approaches  \cite{Fan:2018:UPR:3282485.3243316, DECAMEL, Lin2019ABC} rely on clustering or transferring knowledge by using pseudo labels from a labelled source data to unlabeled target data.  However, different identities may have the same pseudo label, which can make it hard for the model to distinguish similar people.

\subsection{Pose Variation and re-ID}
Person re-ID is inhibited by variation in poses across camera networks. The same individual can appear to be a different person when they are captured with a different pose. This affects the appearance and hence the identity of a person which makes re-ID a challenging task.

Many previous studies have sought to address the impact of pose variation on re-ID performance. \cite{DPFL,Liao_2015_CVPR,8237527,8099654,6619307,DNS} introduced hand-crafted methods such as estimating the body orientation using a histogram of oriented gradients (HOG) detector; or using front, profile and back pose types to extract features following the pose variation. Recently, the adverse effects of a person's pose changes on deep learning based person re-ID models has been recognised and researchers have sought to address this. \cite{su2017pose} first divides the human body into six body parts and fourteen body joints. Features are then extracted from the six local body parts. The global and local body part features are concatenated to get the final representation which is used for re-ID. The same idea was also applied in \cite{zhao2017spindle} where fourteen body joints are detected from the whole body which are then assigned to seven body sub regions such as the head-shoulder, upper body, lower body, two arms, and two leg regions. At the feature fusion stage, all the features extracted from the seven body parts are combined to get a final feature vector. In \cite{Zhao_2017_ICCV}, human body parts are also detected to compute representations over 
the body parts, the calculated similarities are then combined.  \cite{Zhao_2017_ICCV} argue that their approach is more robust to human pose changes as they perform human body part partition. 

More recent re-ID works \cite{ge2018fd,ZHANG202022,Liu_2018_CVPR,10.1007/978-3-030-01240-3_40,saquib2018pose} adopted generative adversarial networks to generate new images with various poses. Both \cite{ZHANG202022} and \cite{Liu_2018_CVPR} adopted the method proposed in Cao et al. \cite{DBLP:journals/corr/abs-1812-08008} to generate the pose skeletons from the MARS dataset. The work in \cite{Liu_2018_CVPR} is inspired by the conditional GAN \cite{DBLP:journals/corr/MirzaO14}. The  generated samples with pose variations are fed into the re-ID network along-with the real images and a guider sub-network is used with the triplet loss to force intra-class features to lie close to each other.  In triplet selection, the positive images are chosen from the real images while  the  negative images are taken from the generated samples to enhance diversity in the training data. \cite{ZHANG202022} also adopted a conditional GAN based network to generate images from the pose skeleton for cross-view person re-ID. PSE \cite{saquib2018pose} takes the front, back and side as a person's coarse pose  and body joint locations as a set of fine pose constraints. To train the CNN with various pose representations, body joint maps are used as inputs along with the original images. However, the above methods require auxiliary pose information in the inference
stage. Ge et al. \cite{ge2018fd} proposed FD-GAN where pose information is no longer required. FD-GAN \cite{ge2018fd} is based on the Siamese architecture and includes two image encoder for two input images, two generators, an identity discriminator to classify whether the input image is real or fake conditioned on the input identity, and a pose discriminator  to classify whether the input image is real or fake conditioned on the target pose. A novel pose loss is also proposed to minimise the difference between
the generated images from the two branches.

In contrast to these approaches, we propose an attention-guided pose transfer network that sequentially learns and transfers the pose to capture the local details of a person while ensuring the preservation of fine image details and the structure of the target pose with the help of an appearance discriminator and pose discriminators respectively. We also ensure to preserve the identity of a person during pose translation using the semantic-consistency loss. Furthermore, for re-ID, we adopted the quartet loss \cite{KHATUN2020102989} to minimise the intra-class distance more than the inter-class distance in feature space.

\section{Proposed Method}
\begin{figure*}
\begin{center}
\includegraphics[width=1.0\linewidth]{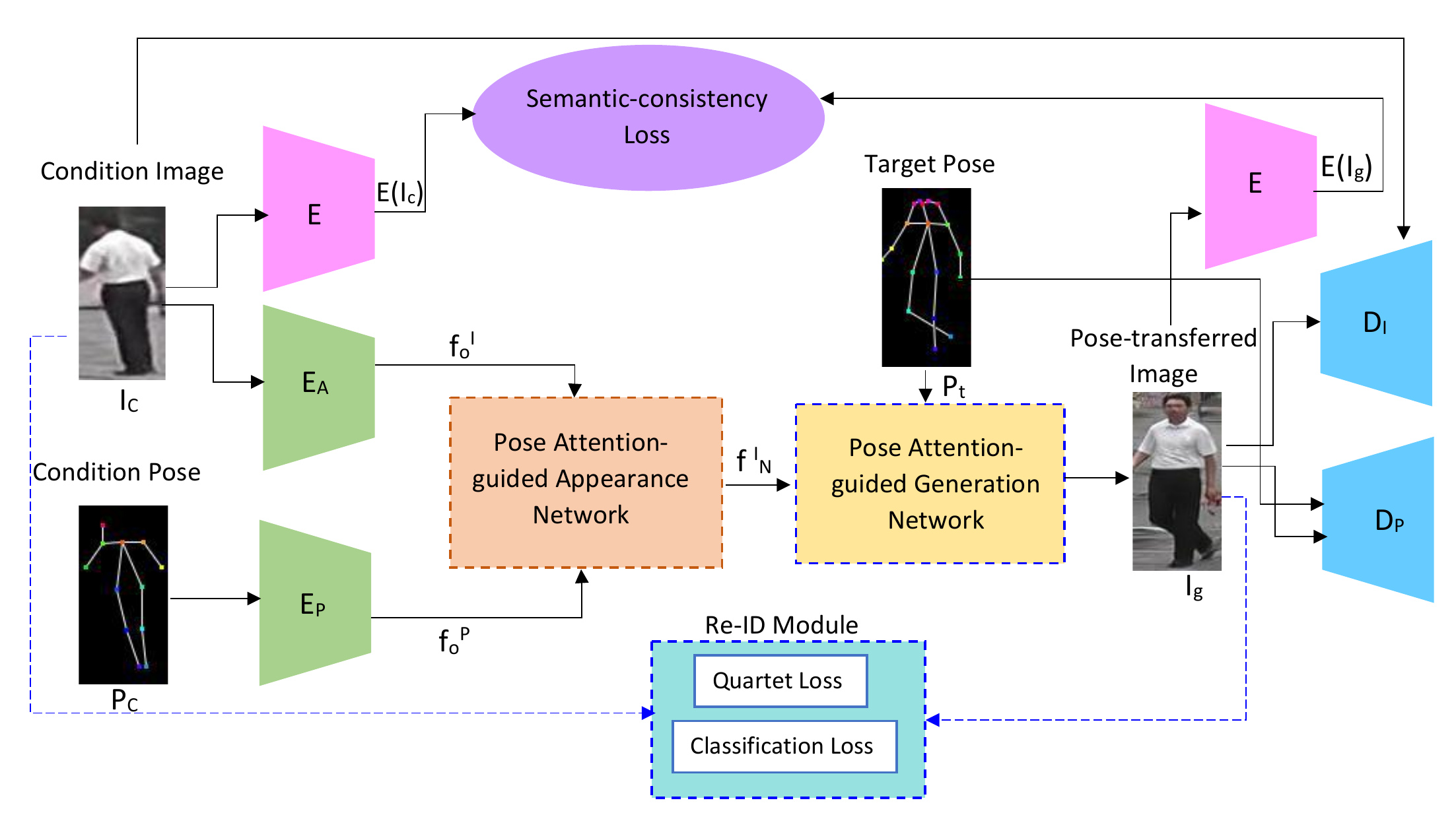}
\end{center}
\caption{Architecture of the proposed pose-guided person image generation and re-ID framework. $E_A, E_P, D_I, D_P$ denote the appearance encoder, pose encoder, appearance discriminator and pose discriminator respectively. The encoder $E$ is used for feature extraction and for the semantic-consistency loss.}
\label{fig:architecture_ch7}
\end{figure*}
The pose attention-guided appearance network and the pose attention-guided generation network progressively model the appearance and shape of a person to synthesise a person image with the target pose, while keeping the appearance and identity constant. The discriminative re-ID module is trained with the quartet loss function \cite{KHATUN2020102989} to boost re-ID performance. As shown in Figure \ref{fig:architecture_ch7}, the condition image, $I_c$, and the condition pose, $P_c$, are first fed into the appearance encoder, $E_A$, and pose encoder, $E_p$, to generate the appearance map, $f_o^I$, and the pose map, $f_o^P$. Both appearance and pose encoders have identical structure and consist of two convolutional layers followed by a batch normalization and ReLU layer. The semantic-consistency loss is applied between the condition image and the pose transferred image to ensure semantic details are preserved between the real and the pose-transferred images. The semantic-consistency loss is applied to the embeddings learnt by the encoder, $E$.

\subsection{Pose Attention-guided Appearance Network}
\begin{figure*}
\begin{center}
\includegraphics[width=1.0\linewidth]{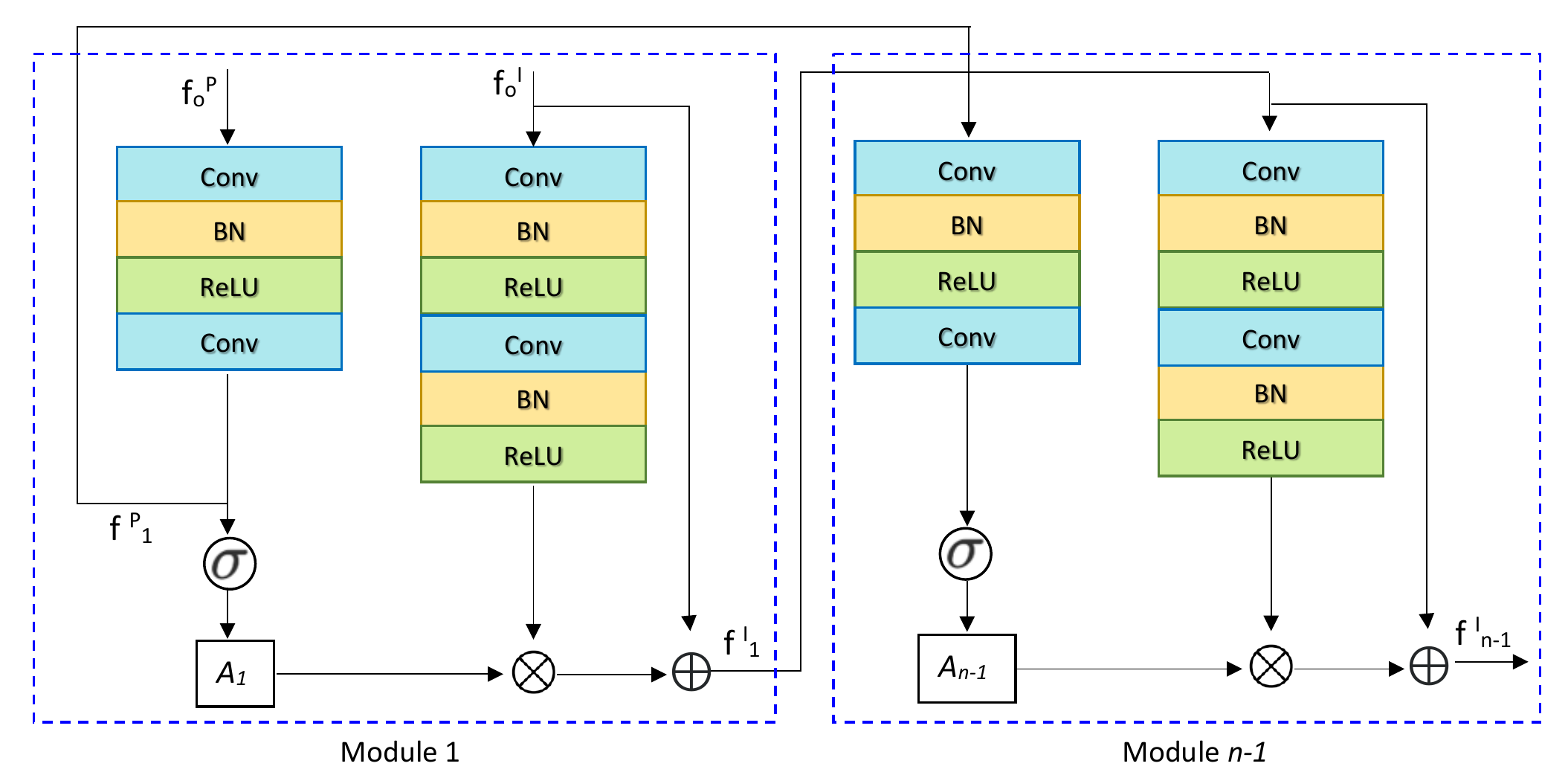}
\end{center}
\caption{Structure of the pose attention-guided appearance network that consists of $N$ blocks. For simplicity, the figure shows two blocks only.  $\sigma$ denotes an element-wise sigmoid function, $\times$ is element-wise multiplication, and $+$ is element-wise addition.}
\label{fig:encoder_ch7}
\end{figure*}

The pose attention-guided appearance network (PAAN) is proposed to learn the appearance representation of a person through a block of attention modules. The initial appearance map, $f_o^I$, and the pose map, $f_o^P$, are fed to the PAAN. The PAAN is illustrated in Figure \ref{fig:encoder_ch7} and consists of  a sequence of $N$ attention-driven encoder modules to progressively learn the appearance representation of a person. There are two streams in each module, each of  which consists of two convolutional layers, followed by a batch normalisation layer and ReLU, and all modules have an identical structure. Considering the first module, the appearance map, $f_o^I$, is passed through two convolutional layers. At the same time, the condition pose map, $f_o^P$, is also passed through two convolution layers to generate the new pose map, $f_1^P$. The new pose map, $f_1^P$, then goes through an element-wise sigmoid activation function to generate the attention mask $A_1$, which ranges between 0 and 1. The attention mask can be represented as,
\begin{align}
A_1 = \sigma(conv(f_o^P)),
\label{eq:1}
\end{align}
where $\sigma$ is an element-wise sigmoid activation function. The attention mask guides the network with respect to where to sample condition patches and directs the appearance stream accordingly.
The transformed appearance map, the output of the convolution layers from the appearance stream, is then multiplied with the attention mask using an element-wise multiplication which is passed through an element-wise addition with the initial appearance map $f_o^I$ to get the new appearance map,
\begin{align}
f_1^I = A_1 \otimes(conv(f_o^I)) \oplus f_o^I,
\label{eq:1}
\end{align}
where $\otimes$ is an element-wise multiplication operation. The resultant appearance map, $f_1^I$, and the pose map, $f_1^P$, are the inputs for the next module and so on. Considering the $Nth$ module, the final outputs of the pose attention-guided appearance network are $f_N^I$ and $f_N^P$.

\subsection{Pose Attention-guided Generation Network}
\begin{figure*}
\begin{center}
\includegraphics[width=1.0\linewidth]{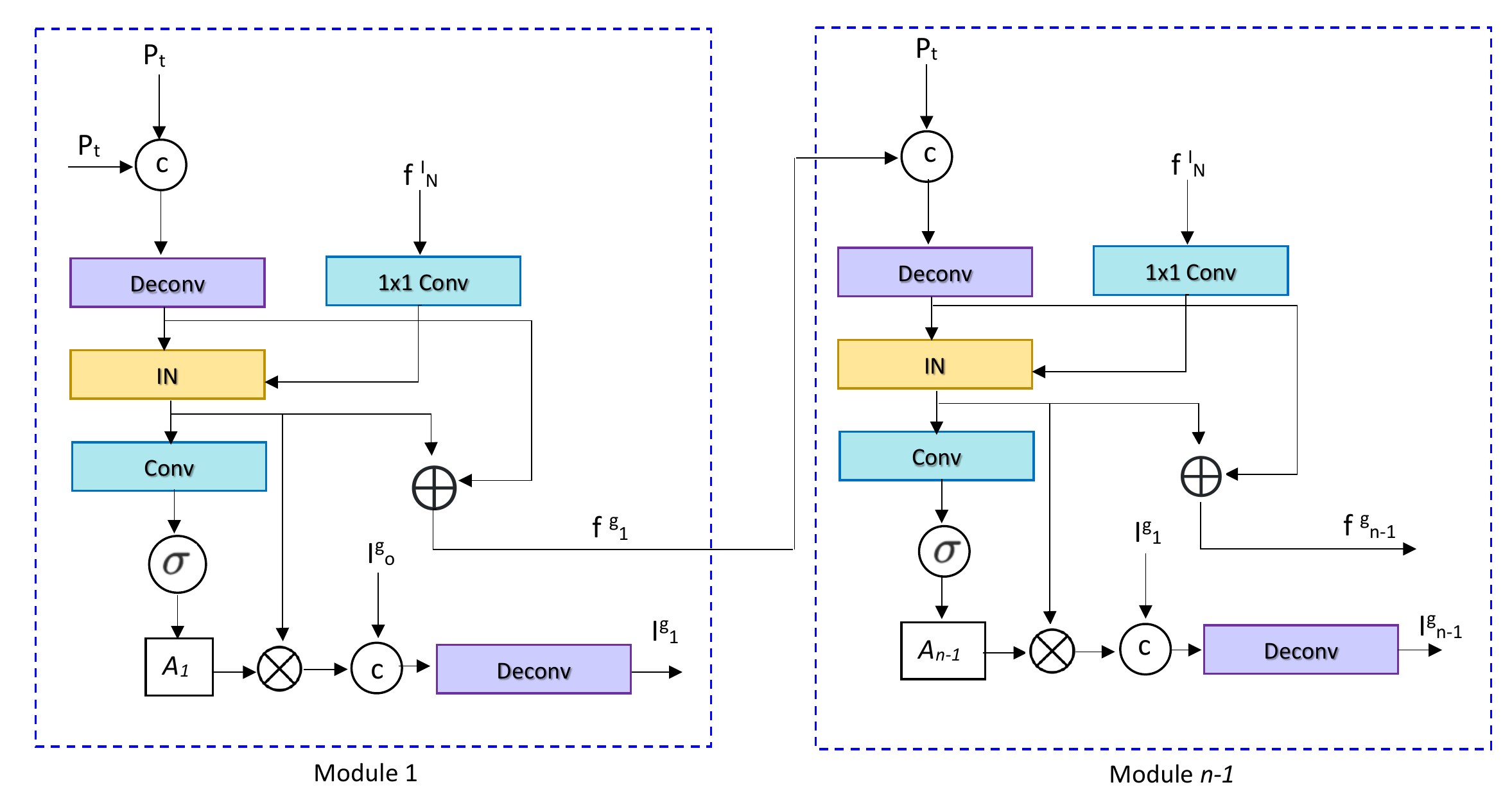}
\end{center}
\caption{Structure of the pose attention-guided image generation network that consists of $N$ blocks. For simplicity, the figure shows two blocks only. $c$ represent channel-wise concatenation, $\sigma$ denotes element-wise sigmoid function, $\times$ is element-wise multiplication, and $+$ is an element-wise addition.}
\label{fig:decoder_ch7}
\end{figure*}
Our pose attention-guided generation network (PAGN) generates the final pose-transferred image under the guidance of the appearance map and consists of $N$ modules, as per the appearance network. The modules are illustrated in Figure \ref{fig:decoder_ch7}. Each module consists of two streams, one is the appearance stream and the other is the image stream. The new appearance map is generated under the guidance of the target pose map, $P_t$, and the appearance map from the PAAN. More specifically, considering the $n-1 th$ module, a channel-wise concatenation operation is performed between the target pose map, $P_t$, and the generator's previous appearance map, $f_1^g$, and then passed through a deconvolutional layer. The final appearance map from the pose attention-guided appearance network (PAAN), $f_N^I$, is fed to a 1x1 convolutional layer. Following \cite{karras2019style}, we adopt adaptive instance normalisation to progressively combine the target pose map, $P_t$, with the appearance map, $f_N^I$. Adaptive instance normalisation requires the style and content, which are the pose and appearance respectively in our case. Thus, the output of adaptive instance normalisation can be represented as,
\begin{align}
x = AdaIN(\underbrace{deconv(f_1^g || P_t)}_\text{style, a}, \underbrace{conv(f_N^I)}_\text{content, b}),
\label{eq:1}
\end{align} 
where $||$ denotes the channel-wise concatenation operation and $conv$ is a 1x1 convolutional layer. The first term is the style while the second is the content.

The final appearance map for this module can be obtained by performing a sum operation between the normalised feature, $x$, is and the style, $a$, and can be represented as,
\begin{align}
f^g_{n-1} = x + deconv(f_1^g || P_t).
\label{eq:1}
\end{align} 
To tell the network where to put the target patches, the normalised output, $x$, is first fed into a convolutional layer with a normalization layer and ReLU, followed by an element-wise sigmoid activation function to get the attention mask,
\begin{align}
A_{n-1} = \sigma(conv(x)).
\label{eq:1}
\end{align}

Finally, the feature maps for the generated image of module $n-1$ are,
\begin{align}
I^g_{n-1} = deconv(I^g_{1} || (A_{n-1} \otimes x)),
\label{eq:1}
\end{align}
where $\otimes$ is an element-wise multiplication between the attention mask and the output of the adaptive normalisation, which is then concatenated with the generated image stream from the last module using channel-wise concatenation. To generate the final feature map of the pose-transformed image for this block, $I^g_{n-1}$ a deconvolution operation is performed. Thus, the outputs of module $n-1$, $f^g_{n-1}$ and $I^g_{n-1}$, are the inputs to the last module, $N$, along-with the target pose map, $P_t$. Assuming we have N generator modules, the final output will be $I^g_{N}$ and $f^g_{N}$, which are then concatenated to get the final  pose-transferred image, $I^g$, which retains appearance and identity information.

\subsection{Appearance Discriminator}
We aim to transfer the target pose of the person in the source image while preserving their identity. Thus, an appearance discriminator, $D_I$, is proposed to ensure that fine image details are realistic in the newly generated pose-transferred images. Hence, the generated pose-transferred image, $I_g$, and the condition image, $I_c$, are fed into the discriminator, $D_I$, to obtain the adversarial loss for appearance discriminator, $D_I$. This can be formulated as,
\begin{align}
L_{GAN}^I = E [log D_I(I_c, I_t) + log(1- D_I(I_c, I_g))].
\label{eq:1}
\end{align}
\subsection{Pose Discriminator}
A pose discriminator, $D_P$, is proposed to identify whether the generated person image, $I_g$, has the same pose as in the target pose map, $P_t$. Thus, the generated image and the target pose maps are first concatenated along the depth dimension and then fed into the convolution, batch normalisation and ReLU layers. A sigmoid operation is then performed to obtain the matching score. The adversarial loss for the pose discriminator can be represented as,
\begin{align}
L_{GAN}^P = E [log D_P(P_t, I_t) + log(1- D_I(P_t, I_g))].
\label{eq:1}
\end{align}

\subsection{Reconstruction loss}
The job of the generator is not only to fool the discriminators, but also to produce images which are close to the real target images. Thus, the $L_1$ pixel-level loss is used to compute the $L_1$ difference between the generated images and corresponding real target images, which helps the generator to achieve more robust convergence during training. The reconstruction loss is written as,
\begin{align}
L_R = ||I_g - I_g^{\prime}||_1,
\label{eq:1}
\end{align}
where $I_g$ are the generated images and $I_g^{\prime}$ are the corresponding ground-truth target images. 

\subsection{Semantic-consistency loss}
In order to preserve the semantic details, we propose to use a semantic-consistency loss at the feature-level to ensure that information critical to re-id is correctly mapped. The semantic-consistency loss is represented as,
\begin{align}
L_{S} = ||E(I_c) - E(I_g)||_1.
\label{eq:4}
\end{align}
To preserve the same semantic details between the condition image the and the pose-translated images, this semantic-consistency loss is used on the learned embeddings extracted by the encoder. Here, $E(I_g)$ represents the embedding of the translated images and $E(I_c)$ are the embeddings of the real input images.

%\subsection{Overall Training Objective}
The objective of our proposed image generation method thus becomes:
\begin{align}
L_{total}= \lambda_{GAN}(L_{GAN}^I +L_{GAN}^P)+\lambda_{R}L_R+\lambda_{S}L_S, 
\label{eq:7.11}
\end{align}
where $\lambda_{GAN}$, $\lambda_{R}$, $\lambda_{S}$ are the weighting factors for the image generation tasks.

\subsection{Discriminative Re-ID Network}
For re-identification, we adopt the improved quartet loss function for verification \cite{KHATUN2020102989} that takes four images as input. For the improved quartet loss, the input images are represented as, $I_i$ = $I_i^1$, $I_i^2$, $I_i^3$, $I_i^4$ where $I_i^1$ is the anchor image, $I_i^2$ is the positive image, and $I_i^3$ and $I_i^4$ are two different negative images. The widely adopted triplet loss pushes images of the same identity close to each other only when the probe images come from the same identity. However, this is impractical in the real-world as in real-world scenarios, where the target data is totally unseen. It is also not specified in the triplet how close the positive pair should be in feature space. Thus features from the same identity may be close to each other, but with large intra-class distances, that leads to a re-ID performance drop. To tackle these issues, the improved quartet loss proposed in \cite{KHATUN2020102989} seeks to minimise the distance between the positive pair over the distance between the negative pair, regardless of whether the probe image belongs to the same person or not and at the same time inserts an additional term to further ensure the intra-class distance is less than a second margin. The verification loss \cite{KHATUN2020102989}  is represented as,
\begin{align}
L_{quartet} = \sum_{i=1}^{n} \Big(max\big\{\parallel\Theta_w (I_i^1) - \Theta_w (I_i^2) \parallel^2  \nonumber\\ - \parallel \Theta_w (I_i^1) -  \Theta_w (I_i^3) \parallel^2  + \parallel\Theta_w (I_i^1) - \Theta_w (I_i^2)\parallel^2 \nonumber\\ - \parallel\Theta_w (I_i^4) - \Theta_w (I_i^3)  \parallel^2, \tau_1 \big\} \nonumber\\ + max\big\{\parallel\Theta_w (I_i^1) -\Theta_w (I_i^2) \parallel^2, \tau_2 \big\}\Big).
\label{eq:6}
\end{align}
Here, the first term of Equation \ref{eq:6} consists of a positive pair $\Theta_w (I_i^1)$ and $\Theta_w (I_i^2)$ and two negative pairs, ($\Theta_w (I_i^1)$, $\Theta_w (I_i^3)$ and $\Theta_w (I_i^4)$, $\Theta_w (I_i^3)$). The second term forces the distance between positive pair, $\Theta_w (I_i^1)$ and $\Theta_w (I_i^2)$ to be less than the second margin, $\tau_2$, and $\tau_2$ is less than $\tau_1$, ensuring that features for the same identity are close in feature space. 

We further use a softmax classification loss to encourage the feature maps of different identities to be separated in feature space. The classification loss is:
\begin{equation}
L_{id}= -log(p(x)),
\label{eq:7}
\end{equation}
where p(.) is the predicted probability that the input image belongs
to the ground-truth class based on its feature map.

The overall objective of our proposed method thus becomes:
\begin{multline}
L_{total}= \lambda_{GAN}(L_{GAN}^I + L_{GAN}^P)+\lambda_{R}L_R+\lambda_{S}L_S+ \lambda_{quartet}L_{quartet} +\lambda_{id}L_{id}, 
\label{eq:7.11}
\end{multline}
where $\lambda_{quartet}$ and $\lambda_{id}$ are the weights to control the importance of the related loss terms

\begin{figure*}
\begin{center}
\includegraphics[width=1.0\linewidth]{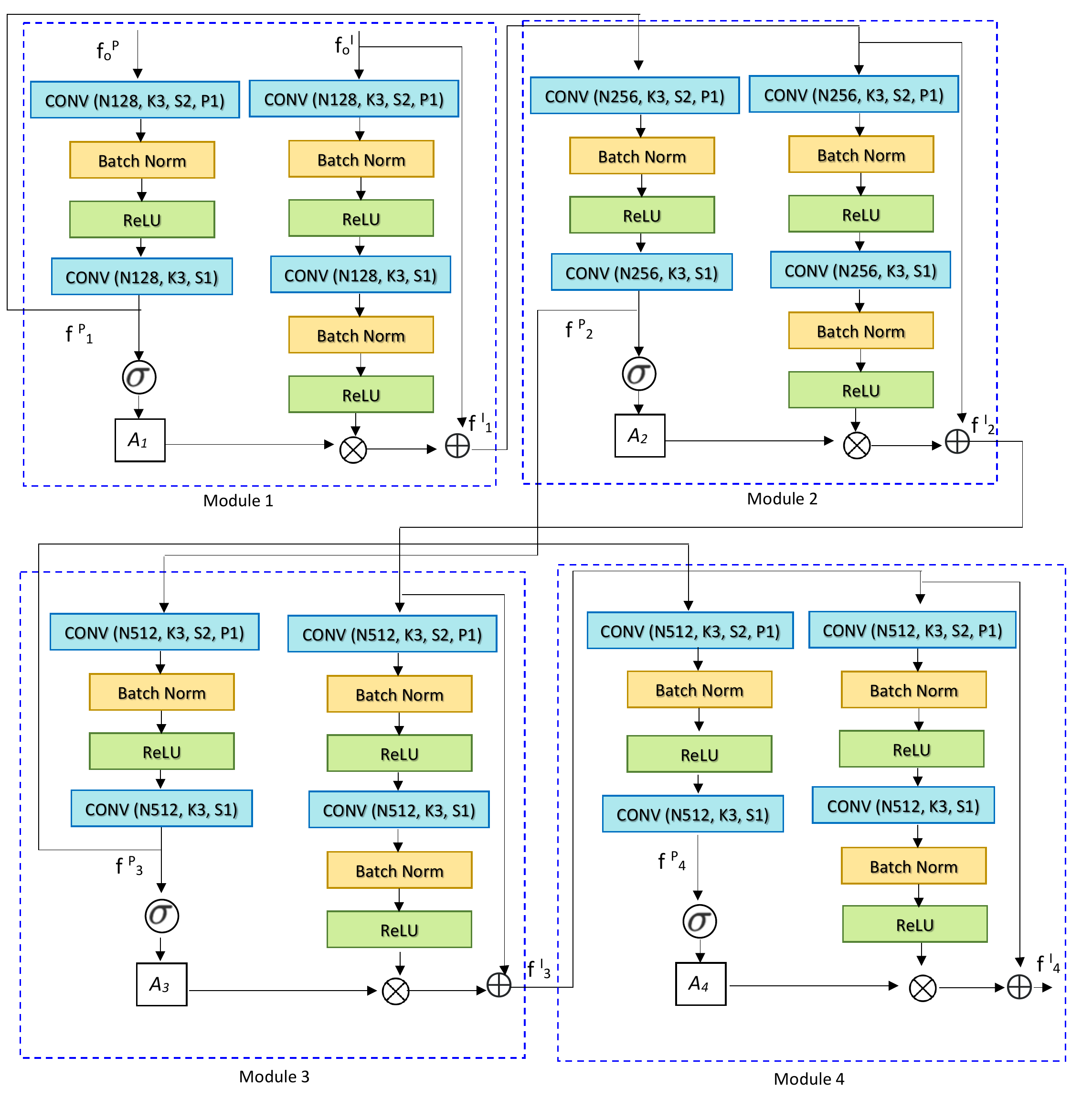}
\end{center}
\caption{Network Structure of the pose attention-guided appearance network (PAAN) with four blocks.}
\label{fig:PAAN_ch7}
\end{figure*}

\section{Network Setup and Training}
We use PyTorch to train our proposed model with the Adam optimizer \cite{adam} for 800 epochs. The learning rate is set to 0.0002 for the first 400 epochs, and then linearly decays towards zero over the next 400 epochs. The weighting factors for $\lambda_{GAN}$, $\lambda_{R}$, $\lambda_{S}$,  $\lambda_{quartet}$ and $\lambda_{id}$ are set to 5, 10, 10, 1 and 1 respectively in Equation \ref{eq:7.11}.
\begin{figure*}
\begin{center}
\includegraphics[width=1.0\linewidth]{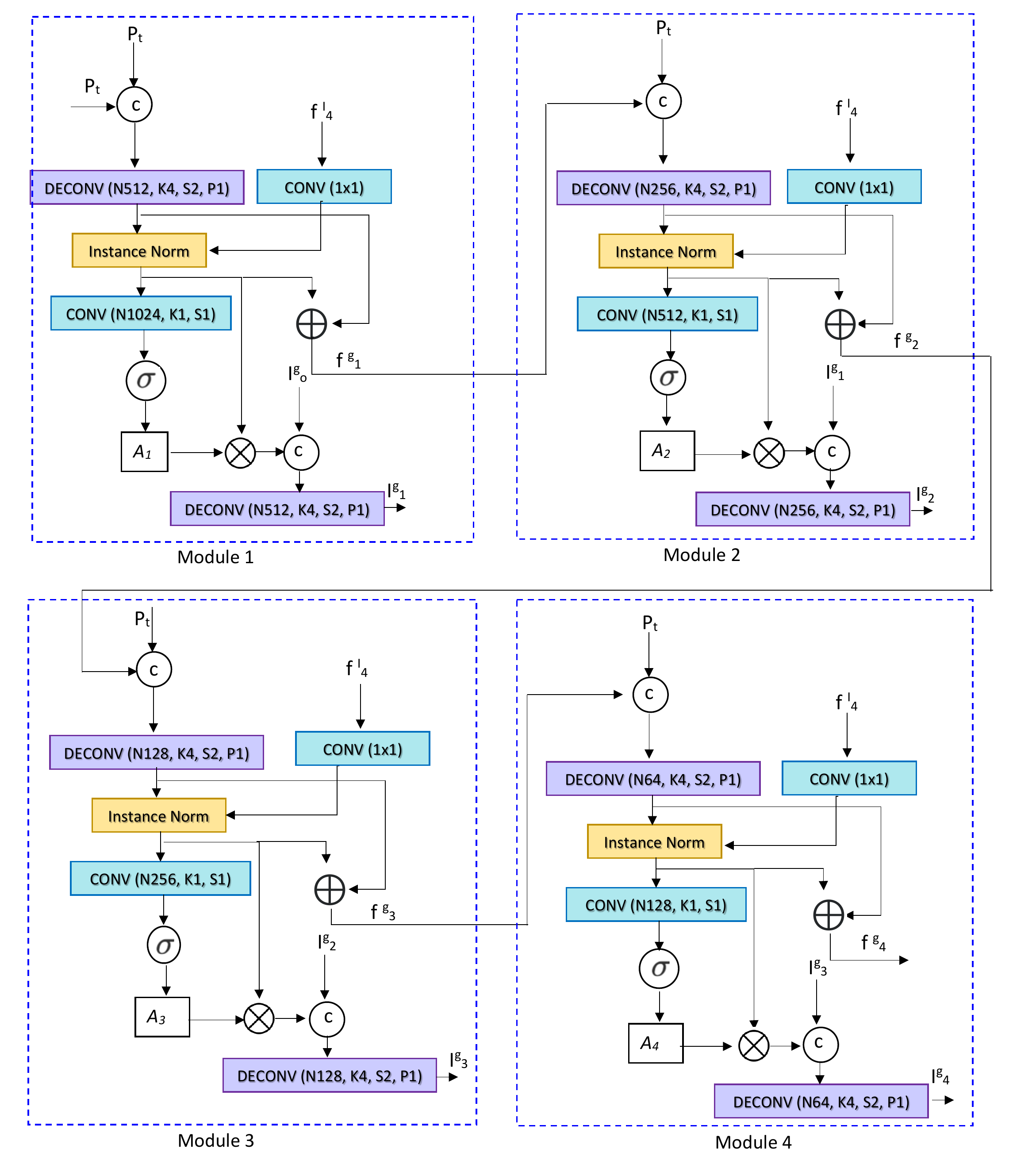}
\end{center}
\caption{Network Structure of the pose attention-guided generation network (PAGN) with four blocks.}
\label{fig:PAGN_ch7}
\end{figure*}

\begin{figure*}
\begin{center}
\includegraphics[width=0.8\linewidth]{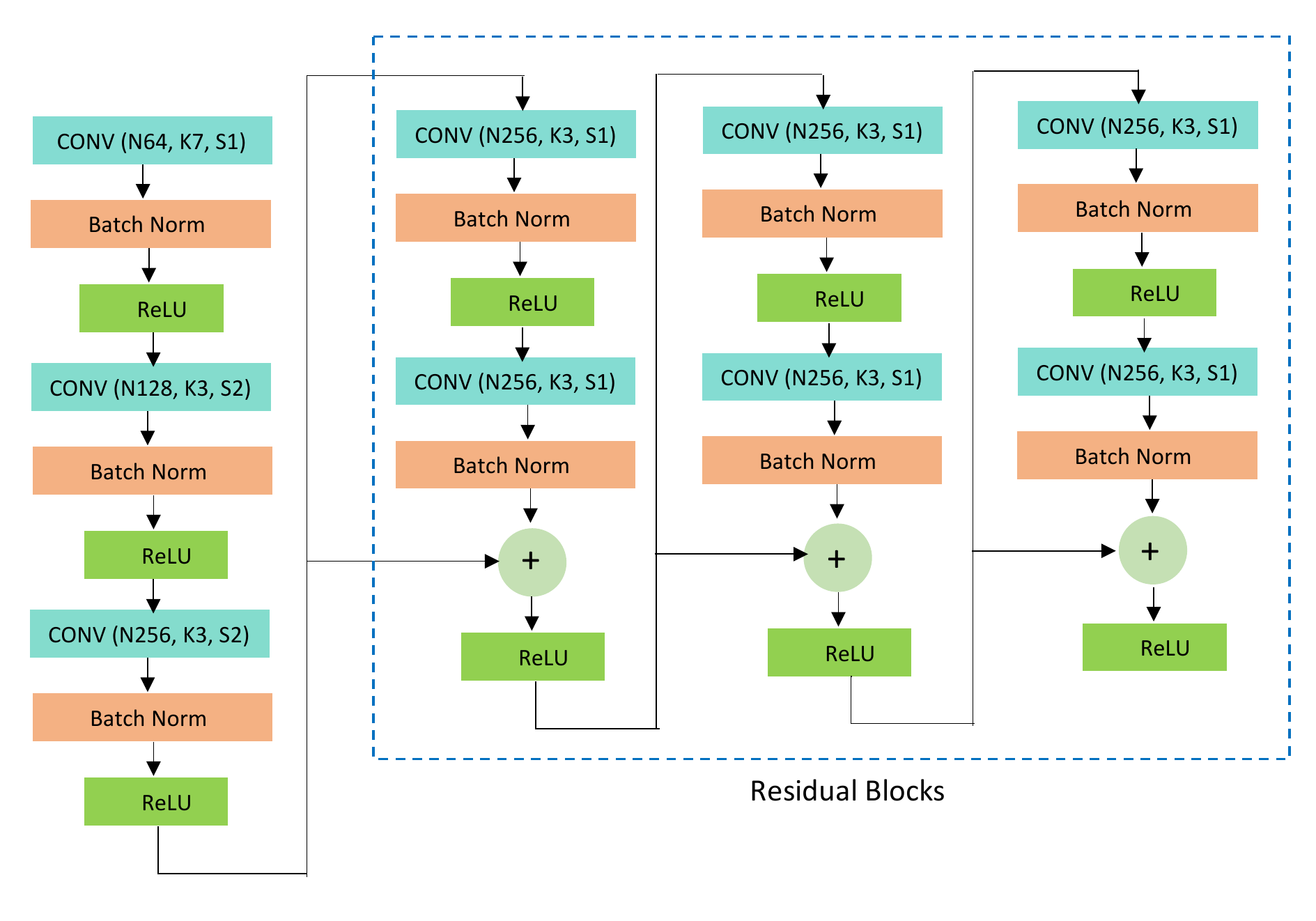}
\end{center}
\caption{Network structure of the discriminator.}
\label{fig:discriminator_ch7}
\end{figure*}

Both the pose attention-guided appearance network (PAAN) and pose attention-guided generation network (PAGN) consists of 4 blocks, each of which consists of two stream as shown in Figure \ref{fig:encoder_ch7} and \ref{fig:decoder_ch7}. The detailed network architecture for the four blocks used by both the PAAN and PAGN are illustrated in Figure \ref{fig:PAAN_ch7} and \ref{fig:PAGN_ch7}. Semantic consistency loss is applied to the embeddings extracted by the encoder.  The network architecture for the appearance and the pose discriminator are identical and are illustrated in Figure \ref{fig:discriminator_ch7}. During image generation, ground-truth target images are given to the discriminator to compute the difference between the generated pose-transferred images and the corresponding real target images as in \cite{ge2018fd}. The classifier takes the feature vector as input, and is a 128-dim fully connected (FC) layer with batch normalization, dropout, and ReLU as the middle layer, and an FC layer with logits equal to the number of identities as the output layer. The dropout rate is set to 0.5 empirically. For the quartet and identification losses, the embedding layer is an FC layer that maps the feature vector to a 128-dim embedding vector.

\section{Experimental results and Discussions}

\subsection {Datasets and evaluation measures}

In this paper, two large and challenging person re-ID datasets, Market-1501 \cite{Market} and DukeMTMC-reID \cite{Duke},  are used for re-ID performance evaluation. We employ OpenPose \cite{DBLP:journals/corr/abs-1812-08008} to locate 18 body joints which are used to generate the source and target pose skeletons which are represented by an 18 channel heat-map. The Market-1501 dataset consists of 12,936  training  images  of 751 identities and 19,732 testing images of 750 identities captured by six cameras in a close to real-world setting. The DukeMTMC-reID dataset consists of 1,404 identities and a total 36,411  hand-drawn bounding boxes extracted from eight high resolution cameras. For the training set, 16,522 images of 702 identities are used and the rest of the identities and images are used for testing. For quartet verification loss, quartets are selected randomly from all the real and generated images, i.e, from the entire augmented training set. Rank-1, rank-5, rank-10 and mAP are used to evaluate the performance of person re-ID.

\subsection{Comparison with state-of-the-art pose-transfer re-ID approaches}
We compare our proposed method with the state-of-the-art pose-guided person re-ID methods, PAC-GAN \cite{ZHANG202022}, Pose-transfer \cite{Liu_2018_CVPR}, PSE \cite{saquib2018pose}, PDC \cite{su2017pose}, PIE \cite{8693885}, Gated Fusion \cite{Bhuiyan_2020_WACV}, PN-GAN \cite{10.1007/978-3-030-01240-3_40}, FD-GAN \cite{ge2018fd} on Market-1501 and Pose-transfer \cite{Liu_2018_CVPR}, PSE \cite{saquib2018pose}, PIE \cite{8693885}, Gated Fusion \cite{Bhuiyan_2020_WACV}, PN-GAN \cite{10.1007/978-3-030-01240-3_40}, FD-GAN \cite{ge2018fd} on the DukeMTMC-reID dataset. The results are reported in Tables \ref{tab:result1_ch7} and \ref{tab:result2_ch7}. 

\begin{table*}
\fontsize{8.0}{8.0}\selectfont
\begin{center}
\begin{tabular}{|p{4.0cm}|p{1.2cm}| p{1.2cm} |p{1.2cm}| p{1.2cm}|}
\hline
\multirow{2}{*}{Method}  & \multicolumn{4}{c|}{Market1501} \\
\cline{2-5}
\cline{2-5}
& Rank1  & Rank5 & Rank10  & mAP \\
\hline
PAC-GAN \cite{ZHANG202022} &75.3 &- &95.7 &-\\
Pose-transfer \cite{Liu_2018_CVPR} &79.8 &- &- &57.9\\
PSE \cite{saquib2018pose} &87.7 &94.5 &96.8 &69.0 \\
PDC \cite{su2017pose} &84.1 &92.7 &94.9 &63.4\\
PIE \cite{8693885} &87.3 &95.6 &86.8 &69.3 \\
PN-GAN \cite{10.1007/978-3-030-01240-3_40} &87.3 &- &- &69.3\\
Gated Fusion \cite{Bhuiyan_2020_WACV} &88.5 &- &- &74.6\\
FD-GAN \cite{ge2018fd} &90.5 &- &- &77.7\\
\hline
%Improved Quartet &91.6 &95.3 &97.4 &75.7\\
\textbf{Pose-guided re-ID (Ours)} &\textbf{93.5} &\textbf{97.1} &\textbf{98.4} &\textbf{78.6}\\
\hline
\end{tabular}
\caption{\label{tab:result1_ch7}Experimental comparison of the proposed approach compared to state-of-the-art methods on Market-1501 dataset. Rank-1, Rank-5, Rank-10 and mAP accuracy are reported.}
\end{center}
\end{table*}

The previous state-of-the-art method is FD-GAN \cite{ge2018fd} for Market1501 with a 90.5\% rank-1 accuracy, where a Siamese architecture is adopted to extract identity-related and pose-invariant features. A novel pose loss is also introduced between generated images from the same identity to further encourage the network to learn identity-related features. For the DukeMTMC-reID dataset, the previous state-of-the-art method was the pose-invariant embedding (PIE) method of \cite{8693885} which achieved 80.8\% rank-1 accuracy and a mAP of 64.1. PIE \cite{8693885} addresses the problem of poor pedestrian alignment and a Pose-Box is introduced to align the pedestrians to a standard pose. To construct the PoseBox, first the human poses are estimated by convolutional pose machines (CPM) to detect the body joints. Affine transformations are then used to project the body parts to a canonical pose. 
\begin{table*}
\fontsize{8.0}{8.0}\selectfont
\begin{center}
\begin{tabular}{|p{4.0cm}|p{1.2cm}| p{1.2cm} |p{1.2cm}| p{1.2cm}|}
\hline
\multirow{2}{*}{Method}  & \multicolumn{4}{c|}{DukeMTMC-reID} \\
\cline{2-5}
\cline{2-5}
& Rank1  & Rank5 & Rank10  & mAP \\
\hline
Pose-transfer \cite{Liu_2018_CVPR} &68.6 &- &- &48.1\\
PSE \cite{saquib2018pose} &79.8 &89.7 &92.2 &62.0 \\
PIE \cite{8693885} &80.8 &88.3 &90.7 &64.1 \\
Gated Fusion \cite{Bhuiyan_2020_WACV} &78.8 &- &- &62.5\\
PN-GAN \cite{10.1007/978-3-030-01240-3_40} &72.8 &- &87.9 &52.5 \\
FD-GAN \cite{ge2018fd} &80.0 &- &- &64.5\\
\hline
%Improved Quartet &82.4 &91.0 &92.5 &77.3\\
\textbf{Pose-guided re-ID (Ours)} &\textbf{84.2} &\textbf{93.6} &\textbf{95.0} &\textbf{78.1}\\
\hline
\end{tabular}
\caption{\label{tab:result2_ch7}Experimental comparison of the proposed approach compared to state-of-the-art methods on DukeMTMC-reID dataset. Rank-1, Rank-5, Rank-10 and mAP accuracy are reported. }
\end{center}
\end{table*}

\begin{figure}
\begin{center}
\includegraphics[width=1.0\linewidth]{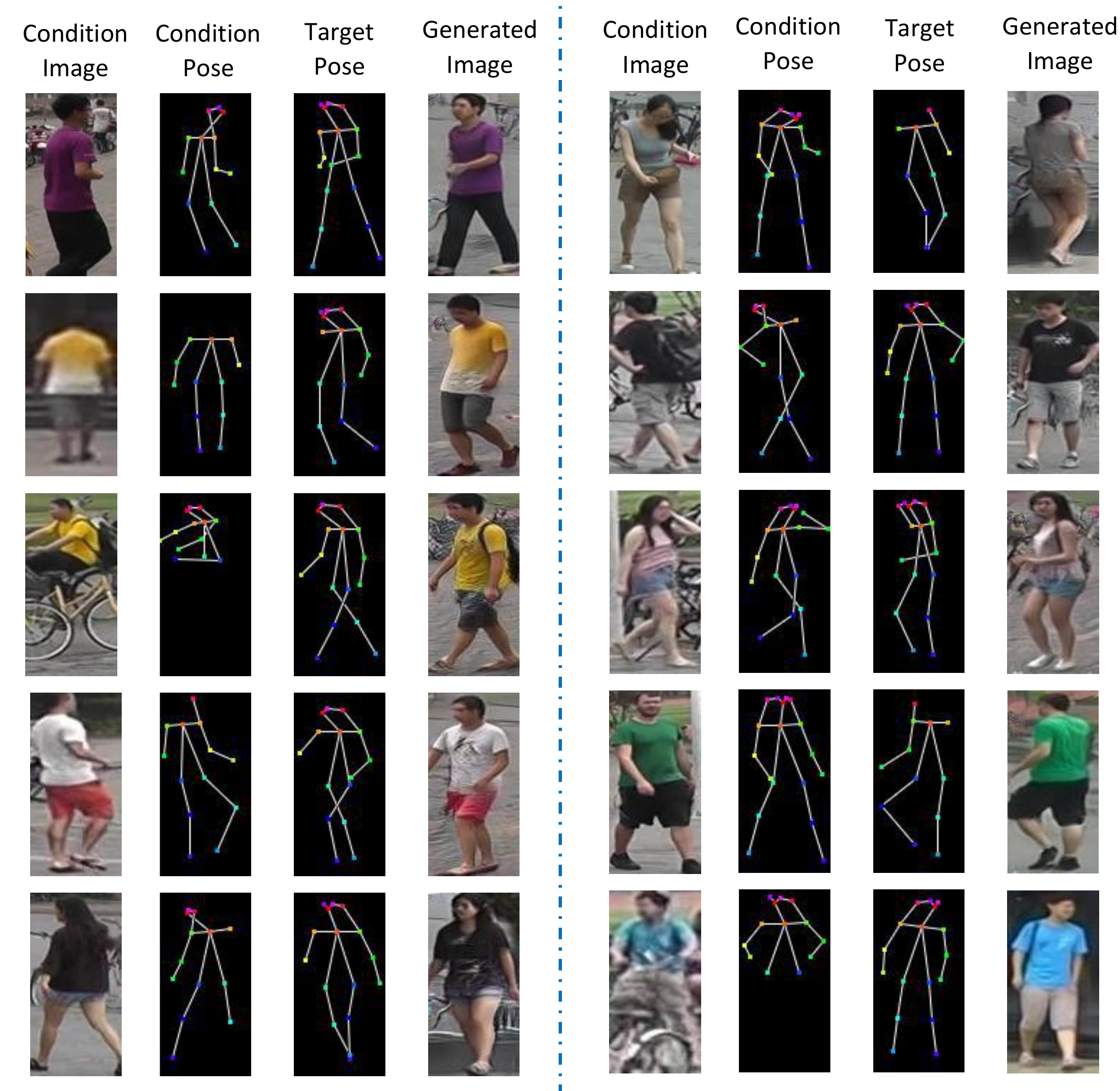}
\end{center}
\caption{Generated samples from Market-1501 dataset. The first column shows the condition images, i.e., the source image, the second, third and fourth column represents the condition pose, the target pose and the generated images respectively.}
\label{fig:generated_market_ch7}
\end{figure}

\begin{figure}
\begin{center}
\includegraphics[width=1.0\linewidth]{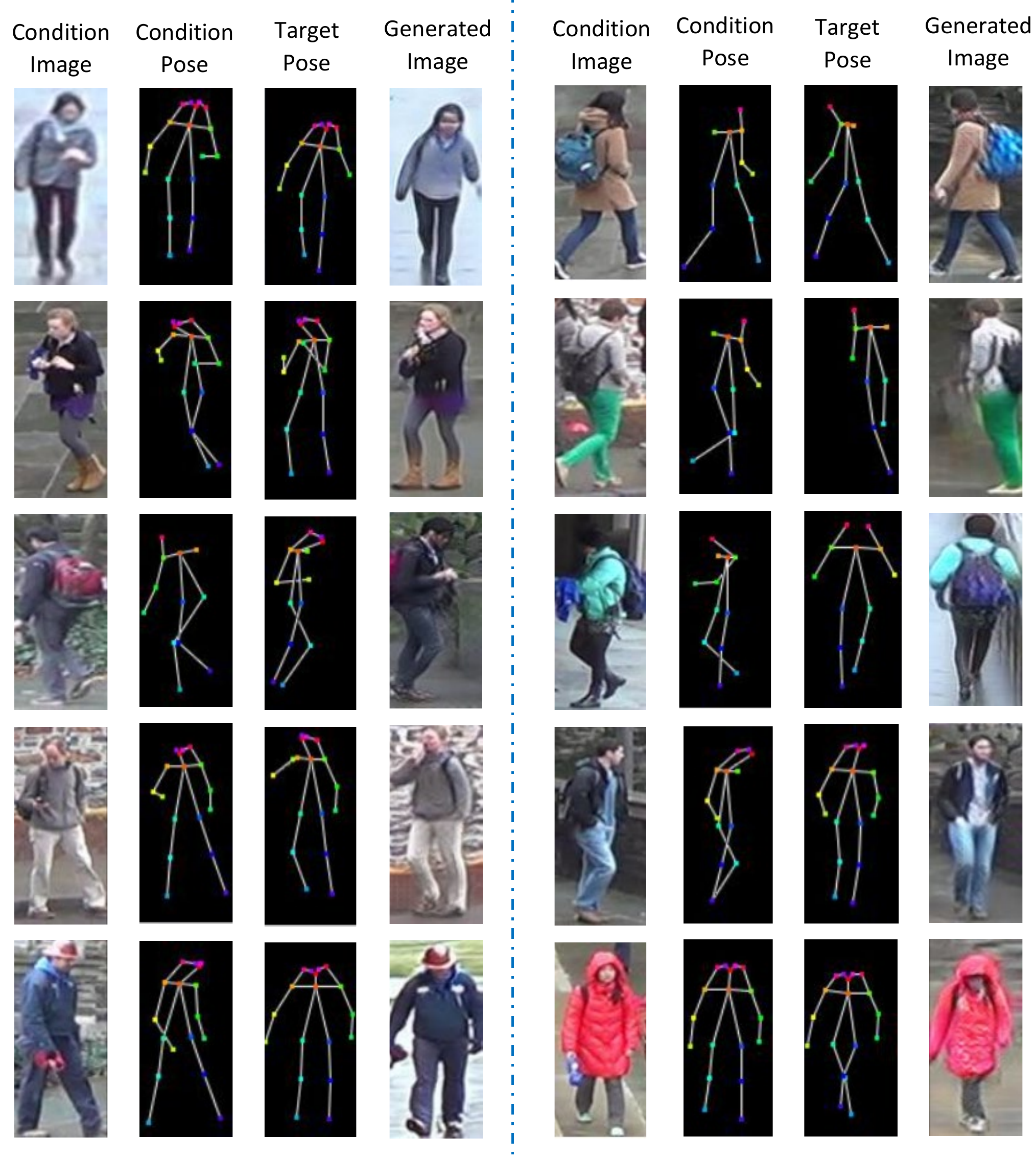}
\end{center}
\caption{Generated samples from DukeMTMC-reID dataset. The first column shows the condition images, i.e., the source image, the second, third and fourth column represents the condition pose, the target pose and the generated images respectively.}
\label{fig:generated_duke_ch7}
\end{figure}

In contrast to these methods, our proposed pose-guided method is different in terms of architecture and loss functions. None of the above-mentioned methods consider capturing local details of a person together with the global structure which is essential to generate person images with new poses while preserving the appearance. The proposed method sequentially learns local details of a person through the blocks of the proposed attention-based appearance network and decodes them in a progressive manner using the attention-based image generation network, moving from small scales to larger scales to combine the local information with the global structure. The proposed method achieves 93.5\% rank-1 accuracy on the Market-1501 dataset which outperforms the state-of-the-art method of \cite{ge2018fd} by 3.0\%. For DukeMTMC-reID, our rank-1 accuracy is 84.2\% while the state-of-the-art method achieved 80.8\%.

Generated pose-transferred samples from Market-1501 and DukeMTMC-reID are shown in Figures \ref{fig:generated_market_ch7} and \ref{fig:generated_duke_ch7}. The first column shows the condition (source) image while the second, third and fourth shows the condition pose, target pose and the generated images. From the figures, it can be seen that the generated images successfully adapt the subject to the target pose while keeping the appearance of the condition images. From Figure  \ref{fig:generated_market_ch7}, we observe that the proposed method successfully transferred to the target pose, even when the source person is riding a bicycle in the source image (see the person wearing yellow t-shirt and riding a bicycle, or the person wearing a blue t-shirt in Figure \ref{fig:generated_market_ch7}). We note that while the DukeMTMC-reID dataset is very challenging due to poor camera resolution, our method is still able to transfer the poses of the condition images according to the target pose map, as illustrated in Figure \ref{fig:generated_duke_ch7}.

\subsection{Qualitative Comparison with Pose-guided Image Generation Approaches}

\begin{figure}
\begin{center}
\includegraphics[width=1.0\linewidth]{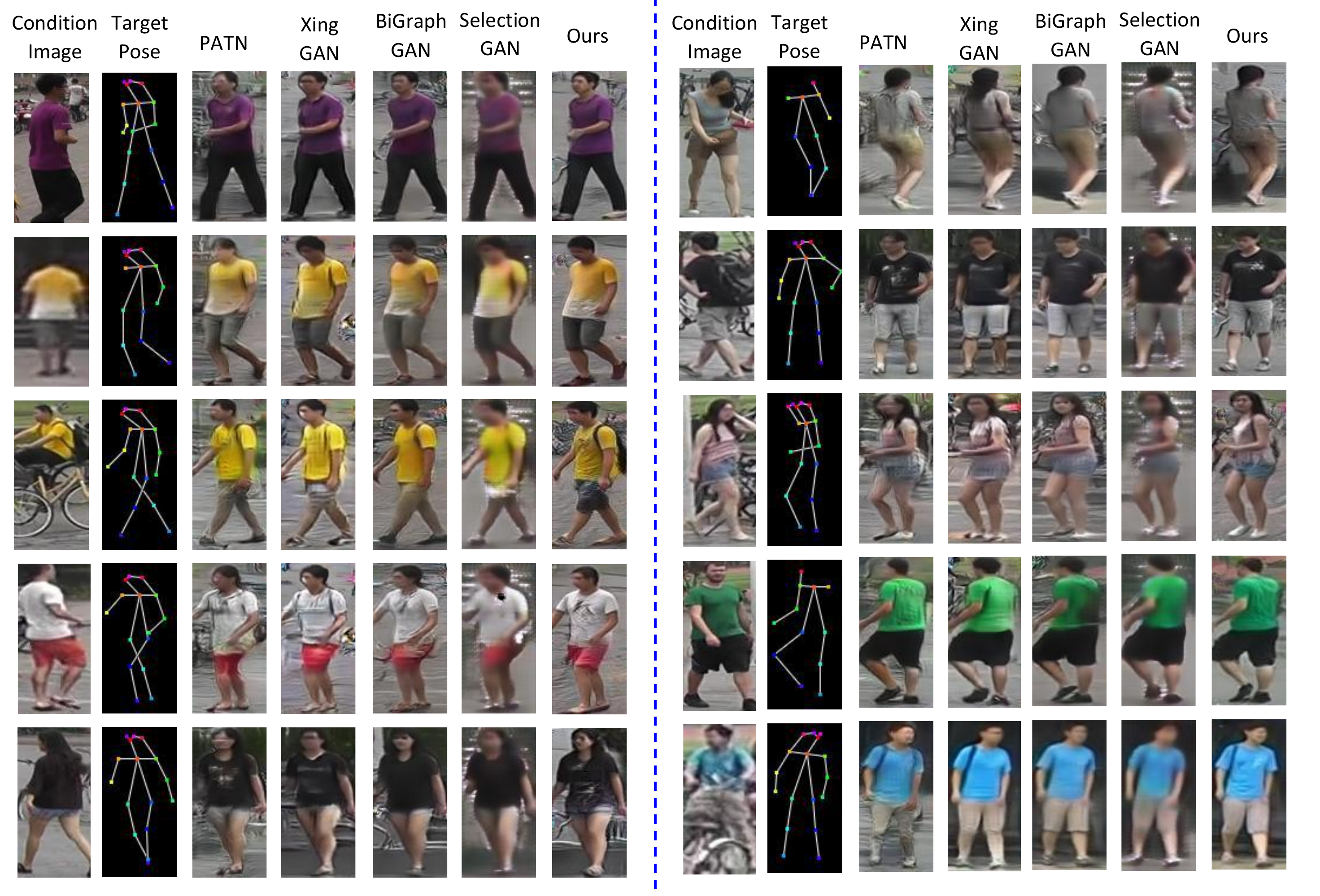}
\end{center}
\caption{Qualitative comparisons of our generated images with the images generated by PATN \cite{zhu2019progressive}, XingGAN \cite{tang2020xinggan}, BiGraphGAN \cite{tang2020bipartite}, and SelectionGAN \cite{tang2020multi} on Market-1501 dataset.}
\label{fig:generated_images_comparison_new}
\end{figure}

A qualitative comparison of state-of-the-art person pose transfer approaches is shown in 
Figure \ref{fig:generated_images_comparison_new} for the Market-1501 dataset. The condition images, target pose maps, and pose-transferred images for PATN \cite{zhu2019progressive}, XingGAN \cite{tang2020xinggan}, BiGraphGAN \cite{tang2020bipartite}, SelectionGAN \cite{tang2020multi} and our proposed method are shown from left to right in Figure \ref{fig:generated_images_comparison_new}. In PATN \cite{zhu2019progressive}, a pose-transfer method is proposed to update the appearance and pose of a person during encoding.  XingGAN \cite{tang2020xinggan} followed \cite{zhu2019progressive} and introduced a cross-conditioning GAN consisting of shape-guided
appearance-based generation and appearance-guided shape-based generation branches to generate more appearance and shape consistent person images. A co-attention fusion module is also proposed to combine the final appearance and shape features. \cite{tang2020bipartite} also extends the approach of \cite{tang2020xinggan} and proposed a bipartite graph reasoning GAN to capture the relations between the condition and target pose through an integration and aggregation module. They also introduce an image fusion module to selectively generate the final result as in \cite{tang2020xinggan}. A multi-channel selection GAN is proposed in \cite{tang2020multi} to improve the quality of the final transferred images by attentively selecting the interested intermediate generations. None of the above-mentioned methods progressively recombine the image in the decoding stage and all these methods are limited to the image generation task. In contrast, we progressively learn and extract the appearance of the condition image during the encoding stage and at the same time progressively transfer the target pose map into the condition image at the decoding stage, thus the local details and the global shape are preserved and combined progressively. From the qualitative evaluation as shown in Figure \ref{fig:generated_images_comparison_new}, our method generates better quality images which are sharper and contain finer details than the pose-transferred images from other state-of-the-art methods. 

\subsection{Ablation Studies}
We perform ablation experiments to evaluate the effectiveness of different numbers of blocks for both  the pose attention-guided appearance network and pose attention-guided generation network. Experiments are also performed to justify the efficacy of the proposed appearance discriminator, the pose discriminator and the semantic-consistency loss. The results are reported in  Table \ref{tab:ablation_ch7}.

\subsubsection{Effectiveness of Number of Blocks for  Appearance and Generation Networks}
\begin{figure}
\begin{center}
\includegraphics[width=1.0\linewidth]{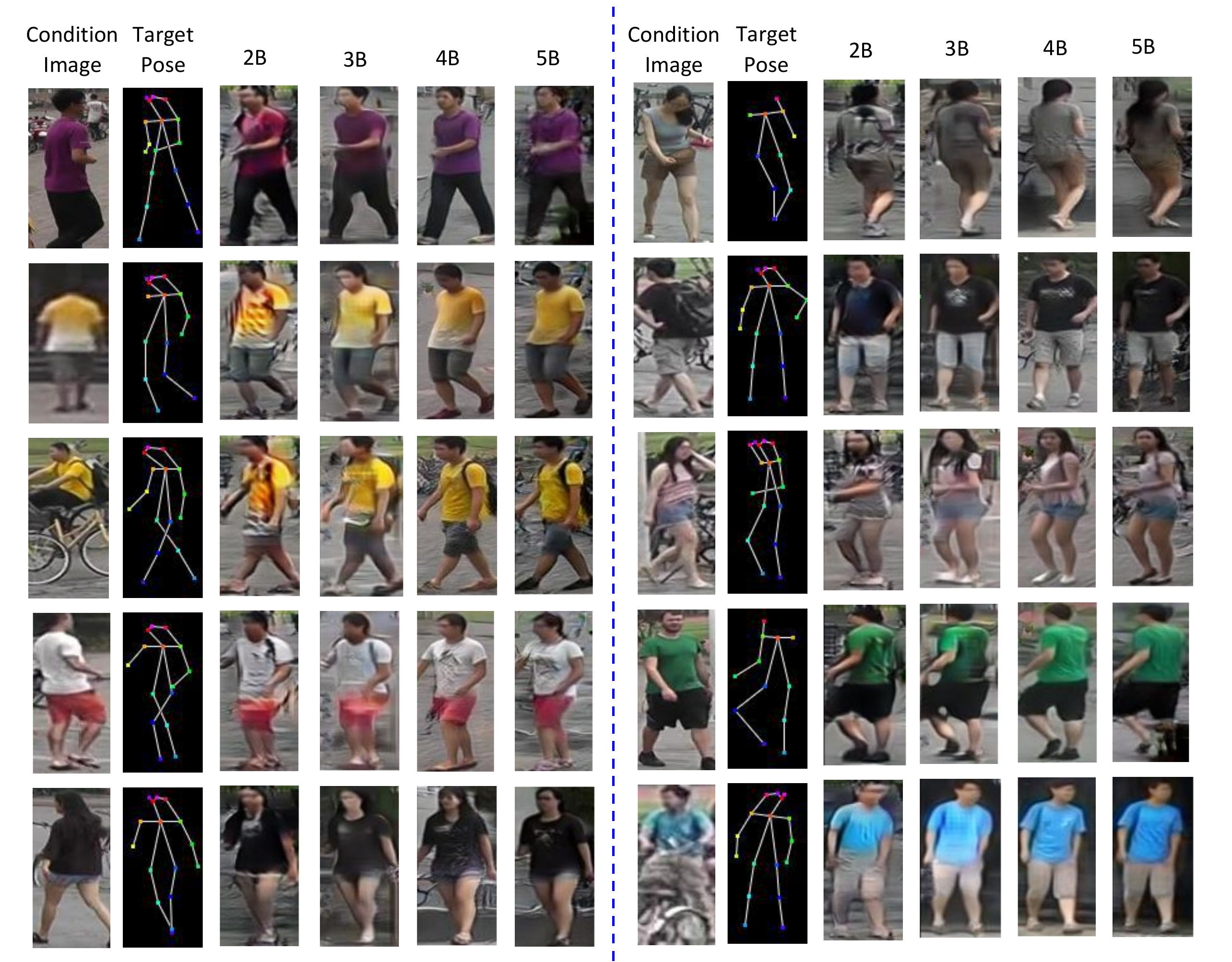}
\end{center}
\caption{Qualitative comparisons of different number of blocks on Market-1501 dataset. 2B, 3B , 4B and 5B represents two, three, four and five number of blocks for both pose attention-guided appearance network and pose attention-guided generation network.}
\label{fig:generated_images_blocks}
\end{figure}

Our pose attention-guided appearance network (PAAN) and pose attention-guided generation network (PAGN) consist of several blocks to learn the appearance and pose of a person, and progressively transfer the target poses based on the condition images. To justify the selection of the number of blocks, we carried out experiments on the Market-1501 dataset to show the effect of differing numbers of blocks. We train the proposed model with two, three, four and five blocks. Qualitative comparisons for different numbers of blocks are shown in Figure \ref{fig:generated_images_blocks}. From Figure \ref{fig:generated_images_blocks}, the pose-transferred images using two blocks have poor quality compared to those generated by three blocks. However, four blocks performs better again and generates images with much finer details. When we further increase the number of blocks to five, we observe that the images produced with both four and five blocks are sharp and lead to pleasant images of a person. 

We further compare the effectiveness of different number of blocks through re-ID performance evaluation. Table \ref{tab:ablation_ch7} shows the rank-1 re-ID accuracy with two, three, four and five PAAN and PAGN blocks. From the results, when considering only two blocks, the rank-1 accuracy is drops by 1.1\% compared to using three blocks, and 4.9\% compared to using four blocks. The same rank-1 accuracy of 93.5\% is achieved for both four and five blocks. As the re-ID performance does not increase further when extending the network to more than four blocks, we select four blocks. From these observations, it is clear that the proposed pose-guided approach only needs a few blocks to capture the local details with the global structure.

\subsubsection{Effectiveness of Appearance Discriminator}
\begin{figure}
\begin{center}
\includegraphics[width=1.0\linewidth]{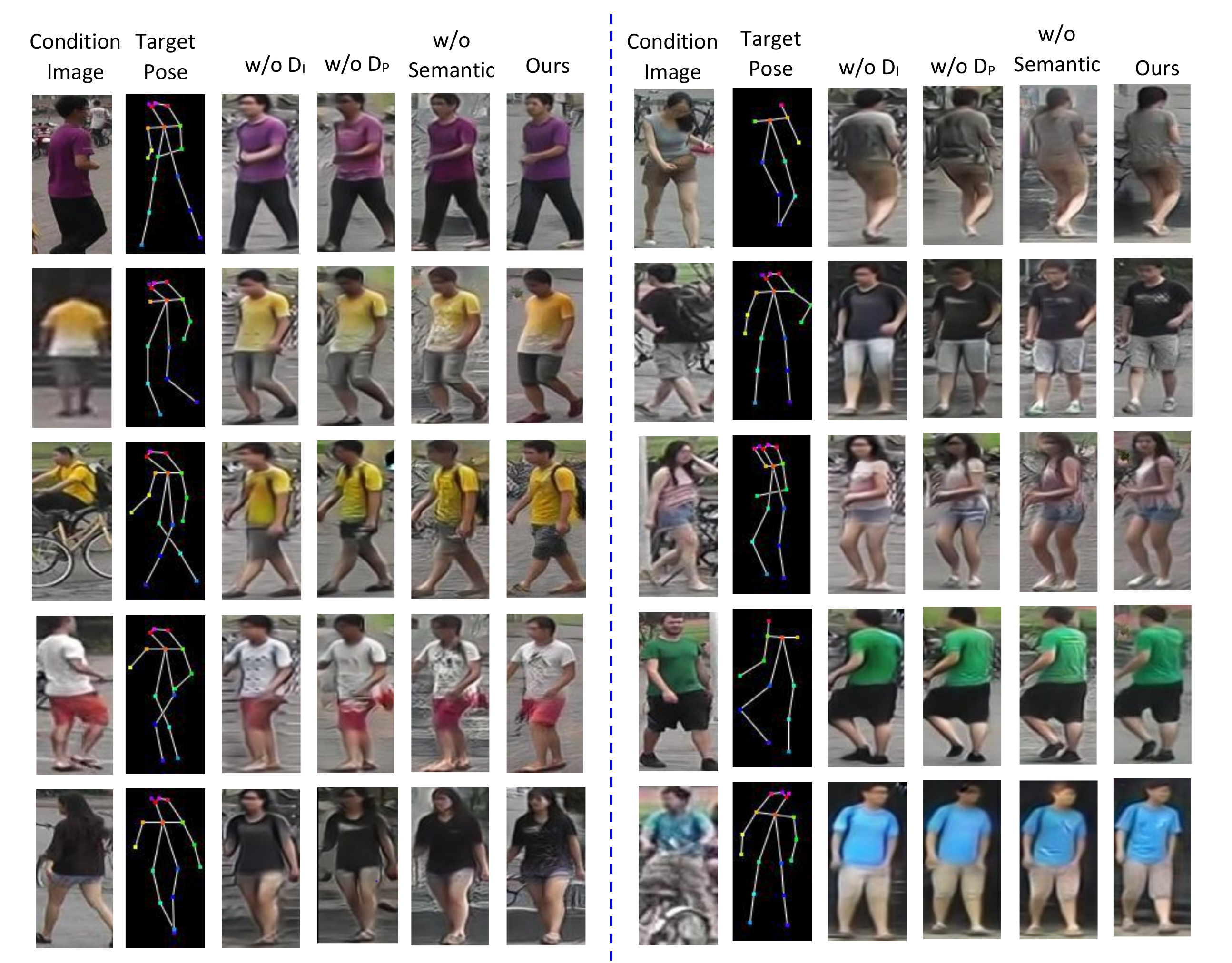}
\end{center}
\caption{Generated pose-transferred samples without appearance discriminator, $D_I$, without pose discriminator $D_P$ and without the semantic-consistency loss are shown in third, fourth and fifth column respectively. The sixth column for both left and right sided images represents the images generated by the complete approach.}
\label{fig:generated_images_discriminators}
\end{figure}

To justify the effectiveness of the proposed appearance discriminator, we perform an experiment when the proposed model is trained with only the pose discriminator and without the appearance discriminator. The results are reported in Table \ref{tab:ablation_ch7}. Removing the appearance discriminator results in a 2.3\% performance drop compared to the complete approach. Figure \ref{fig:generated_images_discriminators} also shows the generated images without the appearance discriminator. From Table \ref{tab:ablation_ch7} and Figure \ref{fig:generated_images_discriminators}, it is clear that without the appearance discriminator, the performance is hampered as the appearance discriminator ensures fine image details are preserved during translation.
\begin{table*}
\fontsize{9.0}{9.0}\selectfont
\begin{center}
\begin{tabular}{|p{9.0cm}|p{1.2cm}|}
\hline
\multirow{2}{*}{Method}  & \multicolumn{1}{c|}{Market1501} \\
& Rank1 \\
\hline
Pose-guided re-ID with two blocks of PAAN and PAGN & 88.6 \\
Pose-guided re-ID with three blocks of PAAN and PAGN & 89.7\\
Pose-guided re-ID with five blocks of PAAN and PAGN & 93.5\\
Pose-guided re-ID without appearance discriminator & 91.2 \\
Pose-guided re-ID without pose discriminator & 89.3 \\
Pose-guided re-ID without semantic-consistency loss & 91.3 \\
\textbf{Pose-guided re-ID} & \textbf{93.5}\\
\hline
\end{tabular}
\caption{\label{tab:ablation_ch7}Ablation studies on Market-1501 dataset in terms of rank-1 accuracy. The model is trained with two, three, four and five blocks of appearance and generation network to compare with the proposed complete approach. Without appearance discriminator: when the model is trained only with the pose discriminator and without pose discriminator: when the model is trained only with the appearance discriminator, and without semantic-consistency loss: when the model is trained without the semantic-consistency loss.}
\end{center}
\end{table*}

\subsubsection{Effectiveness of Pose Discriminator}
We also perform an experiment to evaluate the effectiveness of the proposed pose discriminator which ensures that the pose-transferred images adopt the same structure as the target pose map. In Table \ref{tab:ablation_ch7}, the result is reported as ``pose-guided re-ID without pose discriminator", indicating the model is trained only with the appearance discriminator while ignoring the pose discriminator. There is a 4.2\% drop in rank-1 performance compared to the complete proposed approach. Figure \ref{fig:generated_images_discriminators} also shows the pose-transferred generated images when the model is trained without the pose discriminator.

\subsubsection{Effectiveness of Semantic-consistency Loss}
As the semantic-consistency loss is proposed to ensure that the semantic information is correctly mapped between the real and the generated pose-transferred images, we perform experiment when semantic-consistency loss is not applied to demonstrate how the model performs without the semantic-consistency loss. The result is shown in Table \ref{tab:ablation_ch7} which indicates a 2.2\% drop in performance in rank-1 compared to the complete proposed pose-guided re-ID approach. From Figure \ref{fig:generated_images_discriminators}, it can also be seen that how the semantic-consistency loss affects the generated images.

\section{Conclusion}
In this paper, we propose a pose-guided person image generation network to train a person re-ID model with multiple poses of a person, thus enhancing the discriminative ability of the re-ID model. The proposed model transfers the target pose to the source image without affecting the identity of the person in the source image. During pose transfer where the source image is mapped to the target pose, it is essential to learn the local features along-with the global structure, otherwise it is difficult to transfer complex poses, such as mapping from sitting to standing. To extract and capture local information, a sequential appearance network is proposed. During decoding, we adopt the same sequential approach to progressively recombine local and global features. Additionally, a semantic-consistency loss is imposed between the real and pose-transferred images to ensure that the semantic information is correctly mapped during pose translation and the identity of the person is maintained. Furthermore, we propose to preserve the appearance of the source images during pose translation through an appearance discriminator. As the source image is being mapped to the shape specified by the target pose, a pose discriminator is introduced to ensure pose consistency between the desired and generated poses. Generated images with multiple different poses for each person enhance the performance of person re-identification by increasing pose variations in the training data, leading to state-of-the-art re-identification performance. 

\section*{Acknowledgement} This research was supported by an Australian Research Council (ARC) grant Discovery Grant No: DP200101942.

\bibliography{mybibfile}

\begin{thebibliography}{10}
\expandafter\ifx\csname url\endcsname\relax
  \def\url#1{\texttt{#1}}\fi
\expandafter\ifx\csname urlprefix\endcsname\relax\def\urlprefix{URL }\fi
\expandafter\ifx\csname href\endcsname\relax
  \def\href#1#2{#2} \def\path#1{#1}\fi

\bibitem{DPFL}
Y.~Chen, X.~Zhu, S.~Gong, Person re-identification by deep learning multi-scale
  representations, in: IEEE International Conference on Computer Vision (ICCV),
  2017.

\bibitem{Liao_2015_CVPR}
S.~Liao, Y.~Hu, X.~Zhu, S.~Z. Li, Person re-identification by local maximal
  occurrence representation and metric learning, in: IEEE Conference on
  Computer Vision and Pattern Recognition (CVPR), 2015.

\bibitem{8237527}
J.~{Zhou}, P.~{Yu}, W.~{Tang}, Y.~{Wu}, Efficient online local metric
  adaptation via negative samples for person re-identification, in: IEEE
  International Conference on Computer Vision (ICCV), 2017.

\bibitem{8099654}
S.~Bak, P.~Carr, One-shot metric learning for person re-identification, in: The
  IEEE Conference on Computer Vision and Pattern Recognition (CVPR), 2017.

\bibitem{6619307}
Z.~{Li}, S.~{Chang}, F.~{Liang}, T.~S. {Huang}, L.~{Cao}, J.~R. {Smith},
  Learning locally-adaptive decision functions for person verification, in: The
  IEEE Conference on Computer Vision and Pattern Recognition (CVPR), 2013.

\bibitem{DNS}
L.~Zhang, T.~Xiang, S.~Gong, Learning a discriminative null space for person
  re-identification, in: IEEE conference on computer vision and pattern
  recognition (CVPR), 2016.

\bibitem{su2017pose}
C.~Su, J.~Li, S.~Zhang, J.~Xing, W.~Gao, Q.~Tian, Pose-driven deep
  convolutional model for person re-identification, in: IEEE conference on
  computer vision and pattern recognition (CVPR), 2017.

\bibitem{zhao2017spindle}
H.~Zhao, M.~Tian, S.~Sun, J.~Shao, J.~Yan, S.~Yi, X.~Wang, X.~Tang, Spindle
  net: Person re-identification with human body region guided feature
  decomposition and fusion, The IEEE Conference on Computer Vision and Pattern
  Recognition (CVPR), 2017.

\bibitem{Zhao_2017_ICCV}
L.~Zhao, X.~Li, Y.~Zhuang, J.~Wang, Deeply-learned part-aligned representations
  for person re-identification, in: IEEE International Conference on Computer
  Vision (ICCV), 2017.

\bibitem{ge2018fd}
Y.~Ge, Z.~Li, H.~Zhao, G.~Yin, S.~Yi, X.~Wang, H.~Li, Fd-gan: Pose-guided
  feature distilling gan for robust person re-identification, in: Advances in
  Neural Information Processing Systems (NIPS), 2018.

\bibitem{ZHANG202022}
Pac-gan: An effective pose augmentation scheme for unsupervised cross-view
  person re-identification, Neurocomputing 387 (2020) 22 -- 39.

\bibitem{Liu_2018_CVPR}
J.~Liu, B.~Ni, Y.~Yan, P.~Zhou, S.~Cheng, J.~Hu, Pose transferrable person
  re-identification, in: IEEE Conference on Computer Vision and Pattern
  Recognition (CVPR), 2018.

\bibitem{10.1007/978-3-030-01240-3_40}
X.~Qian, Y.~Fu, T.~Xiang, W.~Wang, J.~Qiu, Y.~Wu, Y.-G. Jiang, X.~Xue,
  Pose-normalized image generation for person re-identification, in: European
  Conference on Computer Vision (ECCV), 2018.

\bibitem{saquib2018pose}
M.~Saquib~Sarfraz, A.~Schumann, A.~Eberle, R.~Stiefelhagen, A pose-sensitive
  embedding for person re-identification with expanded cross neighborhood
  re-ranking, in: IEEE Conference on Computer Vision and Pattern Recognition
  (CVPR), 2018.

\bibitem{NIPS2014_5423}
I.~Goodfellow, J.~Pouget-Abadie, M.~Mirza, B.~Xu, D.~Warde-Farley, S.~Ozair,
  A.~Courville, Y.~Bengio, Generative adversarial nets, in: Advances in Neural
  Information Processing Systems (NIPS), 2014.

\bibitem{DBLP:journals/corr/MirzaO14}
M.~Mirza, S.~Osindero, Conditional generative adversarial nets, CoRR
  abs/1411.1784.

\bibitem{Khatun_2020_WACV}
A.~Khatun, S.~DENMAN, S.~Sridharan, C.~Fookes, Semantic consistency and
  identity mapping multi-component generative adversarial network for person
  re-identification, in: IEEE Winter Conference on Applications of Computer
  Vision (WACV), 2020.

\bibitem{KHATUN2020102989}
A.~Khatun, S.~Denman, S.~Sridharan, C.~Fookes, Joint
  identification–verification for person re-identification: A four stream
  deep learning approach with improved quartet loss function, Computer Vision
  and Image Understanding 197-198 (2020) 102989.

\bibitem{8100115}
P.~Isola, J.~Zhu, T.~Zhou, A.~A. Efros, Image-to-image translation with
  conditional adversarial networks, in: IEEE Conference on Computer Vision and
  Pattern Recognition (CVPR), 2017.

\bibitem{CycleGAN2017}
J.-Y. Zhu, T.~Park, P.~Isola, A.~A. Efros, Unpaired image-to-image translation
  using cycle-consistent adversarial networks, in: IEEE International
  Conference on Computer Vision (ICCV), 2017.

\bibitem{Yi2017DualGANUD}
Z.~Yi, H.~Zhang, P.~Tan, M.~Gong, Dualgan: Unsupervised dual learning for
  image-to-image translation, in: IEEE International Conference on Computer
  Vision (ICCV), 2017.

\bibitem{tang2019cycleincycle}
H.~Tang, D.~Xu, G.~Liu, W.~Wang, N.~Sebe, Y.~Yan, Cycle in cycle generative
  adversarial networks for keypoint-guided image generation, in: ACM Multimedia
  (ACM-MM), 2019.

\bibitem{7780634}
L.~A. Gatys, A.~S. Ecker, M.~Bethge, Image style transfer using convolutional
  neural networks, in: IEEE Conference on Computer Vision and Pattern
  Recognition (CVPR), 2016.

\bibitem{8658643}
M.~M. {Rahman}, C.~{Fookes}, M.~{Baktashmotlagh}, S.~{Sridharan},
  Multi-component image translation for deep domain generalization, in: IEEE
  Winter Conference on Applications of Computer Vision (WACV), 2019.

\bibitem{Yoo2016PixelLevelDT}
D.~Yoo, N.~Kim, S.~Park, A.~S. Paek, I.-S. Kweon, Pixel-level domain transfer,
  in: European Conference on Computer Vision (ECCV), 2016.

\bibitem{8578457}
A.~{Siarohin}, E.~{Sangineto}, S.~{Lathuilière}, N.~{Sebe}, Deformable gans
  for pose-based human image generation, in: IEEE Conference on Computer Vision
  and Pattern Recognition (CVPR), 2018.

\bibitem{tang2020bipartite}
H.~Tang, S.~Bai, P.~H. Torr, N.~Sebe, Bipartite graph reasoning gans for person
  image generation, in: British Machine Vision Conference (BMVC), 2020.

\bibitem{tang2020xinggan}
H.~Tang, S.~Bai, L.~Zhang, P.~H. Torr, N.~Sebe, Xinggan for person image
  generation, in: European Conference on Computer Vision (ECCV), 2020.

\bibitem{zhu2019progressive}
Z.~Zhu, T.~Huang, B.~Shi, M.~Yu, B.~Wang, X.~Bai, Progressive pose attention
  transfer for person image generation, in: IEEE Conference on Computer Vision
  and Pattern Recognition (CVPR), 2019.

\bibitem{DBLP:journals/corr/VariorHW16}
R.~R. Varior, M.~Haloi, G.~Wang, Gated siamese convolutional neural network
  architecture for human re-identification, in: European Conference on Computer
  Vision (ECCV), 2016.

\bibitem{7780513}
F.~Wang, W.~Zuo, L.~Lin, D.~Zhang, L.~Zhang, Joint learning of single-image and
  cross-image representations for person re-identification, in: IEEE Conference
  on Computer Vision and Pattern Recognition (CVPR), 2016.

\bibitem{6909421}
W.~Li, R.~Zhao, T.~Xiao, X.~Wang, Deepreid: Deep filter pairing neural network
  for person re-identification, in: IEEE Conference on Computer Vision and
  Pattern Recognition (CVPR), 2014.

\bibitem{Cheng_2016_CVPR}
D.~Cheng, Y.~Gong, S.~Zhou, J.~Wang, N.~Zheng, Person re-identification by
  multi-channel parts-based cnn with improved triplet loss function, in: IEEE
  Conference on Computer Vision and Pattern Recognition (CVPR), 2016.

\bibitem{DBLP:journals/corr/ChenCZH17}
W.~Chen, X.~Chen, J.~Zhang, K.~Huang, Beyond triplet loss: a deep quadruplet
  network for person re-identification, in: IEEE Conference on Computer Vision
  and Pattern Recognition (CVPR), 2017.

\bibitem{Amena}
A.~Khatun, S.~Denman, S.~Sridharan, C.~Fookes, A deep four-stream siamese
  convolutional neural network with joint verification and identification loss
  for person re-detection, in: IEEE Winter Conference on Applications of
  Computer Vision (WACV), 2018.

\bibitem{Li_2017_CVPR}
D.~Li, X.~Chen, Z.~Zhang, K.~Huang, Learning deep context-aware features over
  body and latent parts for person re-identification, in: IEEE Conference on
  Computer Vision and Pattern Recognition (CVPR), 2017.

\bibitem{98a1e05749b24099a51dcf3c22daefd9}
X.~Chang, T.~M. Hospedales, T.~Xiang, Multi-level factorisation net for person
  re-identification, in: IEEE Conference on Computer Vision and Pattern
  Recognition (CVPR), 2018.

\bibitem{Yang2019PatchBasedDF}
Q.~Yang, H.-X. Yu, A.~Wu, W.-S. Zheng, Patch-based discriminative feature
  learning for unsupervised person re-identification, in: IEEE Conference on
  Computer Vision and Pattern Recognition (CVPR), 2019.

\bibitem{Fan:2018:UPR:3282485.3243316}
H.~Fan, L.~Zheng, C.~Yan, Y.~Yang, Unsupervised person re-identification:
  Clustering and fine-tuning, ACM Transactions on Multimedia Computing,
  Communications, and Applications (TOMM) 14~(4) (2018) 1--18.

\bibitem{DECAMEL}
H.~{Yu}, A.~{Wu}, W.~{Zheng}, Unsupervised person re-identification by deep
  asymmetric metric embedding, IEEE Transactions on Pattern Analysis and
  Machine Intelligence 42~(4) (2020) 956--973.

\bibitem{Lin2019ABC}
Y.~Lin, X.~Dong, L.~Zheng, Y.~Yan, Y.~Yang, A bottom-up clustering approach to
  unsupervised person re-identification, in: Association for the Advancement of
  Artificial Intelligence (AAAI), 2019.

\bibitem{DBLP:journals/corr/abs-1812-08008}
Z.~{Cao}, G.~{Hidalgo}, T.~{Simon}, S.~E. {Wei}, Y.~{Sheikh}, Openpose:
  Realtime multi-person 2d pose estimation using part affinity fields, IEEE
  Transactions on Pattern Analysis and Machine Intelligence (TPAMI) 43~(1)
  (2021) 172--186.

\bibitem{karras2019style}
T.~Karras, S.~Laine, T.~Aila, A style-based generator architecture for
  generative adversarial networks, in: IEEE conference on computer vision and
  pattern recognition (CVPR), 2019.

\bibitem{adam}
D.~Kingma, J.~Ba, Adam: A method for stochastic optimization, in: International
  Conference on Learning Representations (ICLR), 2014.

\bibitem{Market}
L.~Zheng, L.~Shen, L.~Tian, S.~Wang, J.~Wang, Q.~Tian, Scalable person
  re-identification: A benchmark, in: IEEE International Conference on Computer
  Vision (ICCV), 2015.

\bibitem{Duke}
E.~Ristani, F.~Solera, R.~S. Zou, R.~Cucchiara, C.~Tomasi, Performance measures
  and a data set for multi-target, multi-camera tracking, in: European
  Conference on Computer Vision (ECCV) Workshops, 2016.

\bibitem{8693885}
L.~Zheng, Y.~Huang, H.~Lu, Y.~Yang, Pose-invariant embedding for deep person
  re-identification, IEEE Transactions on Image Processing (TIP) 28~(9) (2019)
  4500--4509.

\bibitem{Bhuiyan_2020_WACV}
A.~Bhuiyan, Y.~Liu, P.~Siva, M.~Javan, I.~B. Ayed, E.~Granger, Pose guided
  gated fusion for person re-identification, in: IEEE Winter Conference on
  Applications of Computer Vision (WACV), 2020.

\bibitem{tang2020multi}
H.~Tang, D.~Xu, Y.~Yan, J.~J. Corso, P.~H. Torr, N.~Sebe, Multi-channel
  attention selection gans for guided image-to-image translation, arXiv
  preprint arXiv:2002.01048.

\end{thebibliography}

\end{document}